\documentclass{article}

\usepackage[final]{nips_2018}

\usepackage{graphicx}

\usepackage[utf8]{inputenc} 
\usepackage[T1]{fontenc}    
\usepackage{hyperref}       
\usepackage{url}            
\usepackage{booktabs}       
\usepackage{amsfonts}       
\usepackage{nicefrac}       
\usepackage{microtype}      
\usepackage{color}

\usepackage{enumitem} 
\usepackage{bm}

\title{Dynamic Measurement Scheduling for Adverse Event Forecasting using Deep RL}

\author{
  Chun-Hao Chang*, Mingjie Mai*, Anna Goldenberg \\
  Department of Computer Science, University of Toronto\\
  SickKids Research Institute, Toronto\\
  Vector Institute, Toronto
}

\begin{document}
\maketitle

\vspace{-2mm}

\begin{abstract}
\vspace{-2mm}


Current clinical practice to monitor patients' health follows either regular or heuristic-based lab test (e.g. blood test) scheduling.
Such practice not only gives rise to redundant measurements accruing cost, but may even lead to unnecessary patient discomfort.
From the computational perspective, heuristic-based test scheduling might lead to reduced accuracy of clinical forecasting models. Computationally learning an optimal clinical test scheduling and measurement collection, is likely to lead to both, better predictive models and patient outcome improvement.
We address the scheduling problem using deep reinforcement learning (RL) to achieve high predictive gain and low measurement cost, by scheduling fewer, but strategically timed tests.
We first show that in the simulation our policy outperforms heuristic-based measurement scheduling with higher predictive gain or lower cost measured by accumulated reward. 
We then learn a scheduling policy for mortality forecasting in the real-world clinical dataset (MIMIC3);
our learned policy is able to provide useful clinical insights. 
To our knowledge, this is the first RL application on multi-measurement scheduling problem in the clinical setting.
\vspace{-3mm}





\end{abstract}

\section{Introduction}
Redundant and expensive screening procedures and lab measurements have increased the overall health care costs \citep{feldman2009managing}. 
Recent works also noticed the increasing rate of over-diagnosis \citep{ezzie2007laboratory, moynihan2012preventing, hoffman2012overdiagnosis}.
Numerous studies \citep{iosfina2013implementation, pageler2013embedding} found no evidence that regular blood testing improves diagnosis in hospitals. Frequent blood test may even worsen patient's health \citep{eyster1973nosocomial}.  To combat the situation, \citet{dewan2017reducing} devised a simple heuristic to reduce frequency of blood tests by $87\%$ in pediatric ICU.
Similarly, \citet{kotecha2017reducing} shows that the cost of ordering lab tests can be significantly reduced without increasing in mortality or re-admission rates in cardiac and surgical ICU. These findings point toward the need for principled data-driven approaches for lab test scheduling to improve both the healthcare system  and the patient experience.

Recently  developed time-series forecasting models solve the much needed problem of predicting early detection of adverse events (e.g. sepsis) based on sparse and irregular measurements \citep{ghassemi2015multivariate, soleimani_scalable_2017,futoma_improved_2017}.
However, the timing of these measurements varies from doctor to doctor and from one hospital to another, leading to a drastically different input distribution that may result in inferior classifier performance. Additionally, these classifiers are often not built to provide insight into which measurements help to make the prediction.

We propose to use reinforcement learning (RL) to learn a data-driven and dynamic sampling policy that maximizes a given classifier's performance while minimizing the number and overall cost of needed measurements.
Deep Q-learning is a powerful tool that can be  used to learn from the retrospective data even when the data does not represent optimal behaviors, and has already been shown to be effective for solving clinical problems \citep{raghu2017continuous, futoma2018learning}.
In our work, we first show that in the simulation, when given a near-perfect classifier, our method is able to learn a strategically timed measurement scheduling that outperforms all the heuristic-based scheduling.
Then we test it on MIMIC3, a real Intensive Care Unit (ICU) dataset and compare the learned RL policies to the clinician's policy. 
To our knowledge, this is the first RL application on multi-measurement scheduling problem in the clinical setting.


\begin{figure}
  \vspace{-4mm}

  \includegraphics[width=1.0\textwidth]{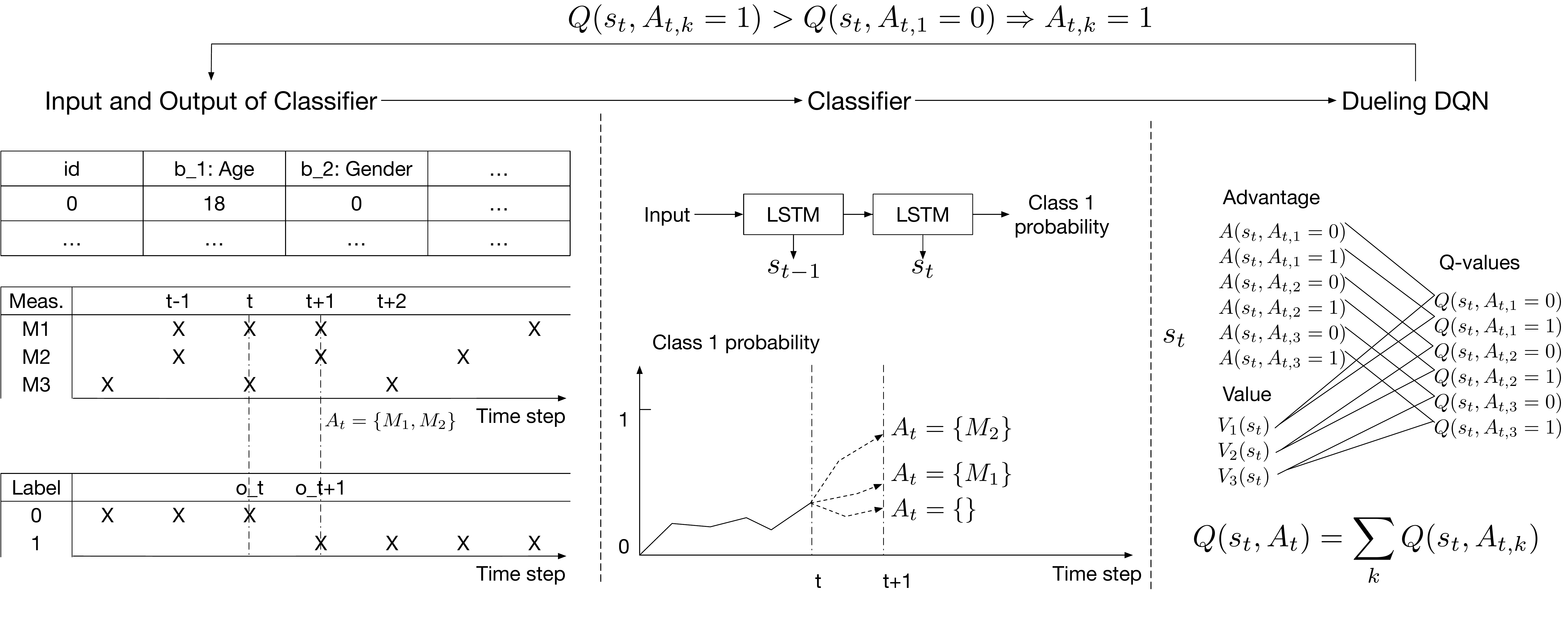}
  \caption{Our System Pipeline. \textbf{Left}: We use mean imputation to impute on evenly spaced grid points for multi-variate, irregularly sampled time series of clinical measurements. \textbf{Center}: Imputed data, missingness indicator and time-invariant features are then feeded into a multi-layer LSTM RNN, which produces an event probability at current time. \textbf{Right}: A dueling DQN agent observes the last LSTM state and learns which time-series measurements should be performed at the next timepoint. 
  To efficiently learn to make multi-action at each time step, we assume the Q-value of each measurement is linearly independent of each other.}
  \label{fig:overall_models}
  \vspace{-8mm}
\end{figure}

\vspace{-3mm}
\section{Methods}
\vspace{-3mm}

Our framework is composed of two parts: a forecasting predictive model and an RL model.
See Figure \ref{fig:overall_models} for an overview.
For the first part, we train an LSTM \citep{hochreiter1997long} classifier to forecast patient’s survival using various multivariate time-series features.
We then train a Dueling deep Q-learning network (DQN) to schedule measurements that maximize the classifier's predictive probability while lowering measurement cost given the history up to the given timepoint.

\textbf{Deep LSTM classifier  }
To handle the sparse time-series data in LSTM, we use mean imputation to fill in the missing measurement values (Figure \ref{fig:overall_models}, Left).
We concatenate the imputed measurement values $\hat{y}^i_t$ with missingness indicators $h^i_t$ and the static demographics $b^i$ for each time step $t$ and individual $i$. 
To learn the classifier, we minimize cross entropy loss between RNN's prediction and true label by backpropogation (Figure \ref{fig:overall_models}, Center). 
All the hyperparameters are listed in appendix \ref{appendix:classifier_performance}.


\textbf{Dueling Deep Q Network  }
Our dueling DQN is adapted from \citet{wang2015dueling}.
We use RNN's last LSTM layer representation $s_t$ as the input to DQN. We represent DQN agent's action $A_t$ using a multi-hot encoding of size $K$, where $A_{t,k}=1$ denotes the $k^{th}$ measurement is going to be observed at the next timepoint and $0$ otherwise.
To handle multi-action at each step, we assume that $Q(s_t, A_t) = \sum_k Q(s_t, A_{t,k})$. Under this assumption, rather than estimating factorial size of Q values for each combination of actions $A_{t, k}$, the DQN network only learns $2K$ Q-values, i.e. $\{Q(s_t, A_{t,k}=0), Q(s_t, A_{t,k}=1)\}_{k=1}^K$ (Figure \ref{fig:overall_models}, Right).  
We optimize the DQN by minimizing the sum of $K$ Bellman-equation square errors, i.e. $\sum_{k=1}^K [Q(s_t, A_{t,k}) - r(s_t, A_{t,k}, s_{t+1}) - \gamma \max_{A_{t+1, k} \in \{0, 1\} } Q(s_{t+1}, A_{t+1, k})]^2$, where $\gamma$ is the reward discounted factor. 
Given an action $A_{t,k}$, the reward $r(s_t, A_{t,k}, s_{t+1})= \lambda g_{t,k} - c_{t, k}$, where $c_{t,k}$ is measurement cost. $g_{t, k} = p(\textrm{event} | A_t=\{M_k\}, o_{t+1}=1) - p(\textrm{event} | A_t = \emptyset, o_{t+1}=1)$ is gain in predictive probability when $o_{t+1}=1$ (e.g. event = 'death within $12$ hours'). And $\lambda$ trades off between gain and cost. 




\vspace{-3mm}
\section{Results}
\vspace{-2mm}

\begin{figure}[t]
  \centering
  \includegraphics[width=1.0\textwidth]{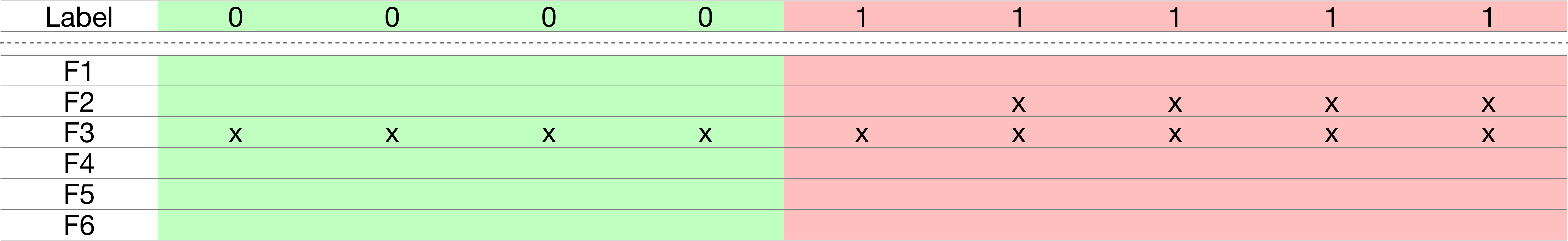}
  \caption{An example trajectory of our DQN policy in the simulation. Our DQN measures constantly on the most important feature F3 and additionally measure F2 when in the critical clinical state $1$.}
  \label{fig:example_policy}
\end{figure}

\subsection{Simulation}
\vspace{-2mm}
The goal of this simulation is to study the performance of the DQN given a near-perfect classifier. Here, we simulate a terminal event forecasting task, used a softmax classifier to produce rewards and then train a DQN agent using the rewards generated by the classifier.

Patient clinical status is simulated to be a binary time series generated under a two-state Markov model: $S = \{0, 1 : 0 = Healthy, 1 = Critical\}$. A consecutive sequence of five $1$s in the status series indicates the onset of a terminal event.
We simulate patients to have different trajectory lengths $T$ indexed by $t$ and $6$ types of input signals indexed by $k$, as follows. Let $\epsilon_{t,k} \sim N(0,1)$ and $c_t\sim Bernoulli(p=0.5)$. The first three types of measurements $(k \in [1,3])$ $y_{t, k} = (1 + \epsilon_{t,k})$ when $S_t=1$ and $(-1 + \epsilon_{t,k})$ otherwise. The last three types of measurements $(k\in[4,6])$ are $\{\epsilon_{t,k}, c_t+\epsilon_{t,k},\epsilon_{t,k}\}$ independent of $S_t$. We randomly remove values from the generated matrix to introduce missingness creating a more realistic scenario. In the case of missingness, the measurement value is set to $0$. 
The measurements are designed such that first three types of measurement (denoted as F1, F2, F3) are of increasing importance while the last three measurements are noise.
We design a classifier considering the feature importance vector $\{f_k\}_{k=1}^6 = $ (4 2 1 0 0 0). 
The classifier takes in time series measurement of the $5$ most recent timepoints $\{\{y_{t,k}\}_{k=1}^6\}_{t=t'-4}^{t'}$, where $t'$ is the current time.
The classifier then forecasts whether the patient is in terminal event within 5 future timepoints with $p(o_t=1) = softmax( \sum_{t=t'-4}^{t'} \sum_{k=1}^6 y_{t, k} \cdot f_k)$.
The classifier increases the certainty of a terminal event when it discovers more critical signals in the measurement values. 
To see whether the agent can discover important features, we employ a uniform action cost $c_{t,k} = 1$.
The RL agent takes $\{\{y_{t,k}\}_{k=1}^6\}_{t=t'-4}^{t'}$ as input. We set reward discount factor $\gamma = 0.99$ and $\lambda = 105$.

We learn a DQN agent on an dataset of $5,000$ patient trajectories simulated as the scheme above. We include several baselines that resemble the heuristic-based test scheduling. These baselines include randomly selecting one of the informative features ('F1-3 random'), selecting only one of the informative features ('F1 alone', 'F2 alone', 'F3 alone'), selecting all the informative features ('F1-3 all'), and selecting any of the two informative features at each time step ('F1-2 alone', 'F2-3 alone'). 

\begin{table}[tbp]
  \centering 
  \caption{Average accumulated reward across $500$ test trajectories.} 
  \begin{tabular}{|l|l|l|l|}\hline
    Policy & Accumulated reward & Policy & Accumulated reward \\\hline
    DQN & $\bm{-19.67}$  & F1-3 random &  $-47.11$ \\\hline 
    F1 alone & $-20.04$ & F1-3 all &  $-84.75$ \\ \hline
    F2 alone &  $-20.59$ &  F1-2 alone & $-50.00$ \\ \hline 
    F3 alone & $-21.12$ & F2-3 alone & $-48.50$ \\ \hline
  \end{tabular}
  \label{tab:acc_reward_simulated} 
  \vspace{-2mm}
\end{table}

We found that the trained DQN agent outperforms other baseline policies in term of average accumulated reward (Table \ref{tab:acc_reward_simulated}). In Figure \ref{fig:example_policy}, we show an example trajectory of our policy. 
It always selects the most informative feature (F3), and additionally selects the second most informative feature (F2) whenever it finds the patient is in a critical state. It doesn't select third most informative feature (F1) or any other noisy features to avoid accruing total measurement cost. 
When learning DQN with $\gamma=0$, the agent is short sighted and not able to discover any critical state by measuring any informative feature. 
It also ends up choosing one of the noisy features deterministically. 

\vspace{-2mm}
\subsection{Results on MIMIC3}
\vspace{-1.5mm}
The goal of our experiment is to schedule measurements to forecast mortality in a real clinical setting. We use $\gamma = 0.95$. Our preprocessing of MIMIC3 is in the Appendix \ref{appendix:preprocessing}. First, we show that we train a well performing RNN classifier: with sufficient information RNN vastly outperforms baselines such as random forest that do not take time into account (Appendix \ref{appendix:classifier_performance}). We then show that combining RNN and DQN, we are able to learn a clinically relevant policy from offline data. 

We investigate two kinds of measurement cost. First, we investigate uniform cost to understand which feature is most important. Second, to simulate a more realistic measurement cost, we set the cost to be inversely proportional to its frequency under physician policy.
Table \ref{tab:top_features} shows the top 5 features selected by physician and our learned DQN agent under two formulation of measurement costs. 
As expected, the majority of the measurements being collected by physician are from regular blood tests, probably due to its static sampling nature. 
Under the uniform costs, the top features being chosen by our DQN agent are clinically relevant to patient's overall health. 
For example, Glasgow Comma Scale total (GCS) represents the level of consciousness, Lactate is a measure of cell-hypoxia in the blood, and Fraction of inspired oxygen (FiO2) indicates quality of breathing.
We find that pH, FiO2 and GCS are highlighted under both formulations of measurement cost, indicating these measurements to be both predictive of mortality and cost-effective. 





\begin{table}[tbp]
  \centering 
  \caption{Top 5 features selected by Physician, DQN with the uniform measurement cost, and DQN with  measurement costs inversely proportional to the physician's measurement frequency.} 
  \begin{tabular}{|l|l|l|l|}\hline
    Rank & Physician  & DQN w/ uniform cost & DQN w/ inverse costs  \\\hline
    1 & Heart Rate & Lactate & Heart Rate \\\hline
    2 & Respiratory rate & Bilirubin & Respiratory rate \\\hline
    3 & Systolic blood pressure & pH & pH \\\hline
    4 & Diastolic blood pressure & Fraction of inspired oxygen & Fraction of inspired oxygen \\\hline
    5 & Mean blood pressure & Glasgow coma scale total & Glasgow coma scale total \\\hline
  \end{tabular}
  \label{tab:top_features}
  \vspace{-3mm}
\end{table}



\vspace{-3mm}
\section{Discussion and Future Work}
\vspace{-3mm}
In this study, we assume ordering lab tests at the current time point gives us the real measurement values at the immediate next time point. 
We can incorporate the time constraint and relax this assumption by delaying the reward that RL agent receives to later time step.
Another concern is that our RL agent might learn to sample features that are overfitting to the classifier. 
To avoid this phenomenon, we plan to scale our work by using an ensemble of classifiers to reduce overfitting and give a robust reward.

Our future work includes giving a more quantitative evaluations using off-policy policy evaluation such as doubly robust method \citep{jiang2015doubly}. We are also planning to relax the linear assumption on Q values to make the Q-learning more unbiased.

\vspace{-3mm}
\section{Related Work} 
\vspace{-3mm}


\textbf{Real-time prediction} Several models have been proposed for real-time prediction on irregularly sampled EHR data \citep{soleimani_scalable_2017, futoma_improved_2017, hwang2017disease, zhang2017medical}. 
Our work is built on top of these predictive models. In addition, we learn what inputs are needed to maximize the utility of classifiers while minimizing overall measurement costs. 

\textbf{Deep RL in health care} Several recent works used RL to learn a treatment plan in ICU \citep{weng2017representation, raghu2017continuous, futoma2018learning}. Our work is a two-tier system that first learns a classifier to represent reward functions and then uses RL similarly but in a space of lab test scheduling.

\textbf{Active feature acquisition}
Several works have been studying the problem of selecting the subset of features and achieve the maximum prediction performance in a non-time-series classifier \citep{contardo2016sequential, he2016active, shim2017pay}. We tackle time-series feature acquisition problem where historical information matters. This is especially true in a health care setting.

\textbf{Active sensing in medical setting}
\citet{ahuja_dpscreen_2017} handles single-measurement scheduling problem for breast cancer screening by adopting a fixed model-based transition model. Unfortunately, it requires strong assumption, knowing the disease model dynamics and does not handle multiple types of measurements.
\citet{yoon2018deep} proposes scheduling measurements to trade between uncertainty in prediction and the measurement cost. 
Our approach differs in three ways. 
First, we use Q-learning with $\gamma > 0$ to learn policy that maximizes cumulative discounted reward of patient trajectories, while they greedily select measurements that would exceed the utility threshold at the next time step.
Second, we consider a different definition of information gain - gain in predictive probability. Consider a binary case, where the model produces a wrong estimate, a measurement that encourages a lower uncertainty would not be the ideal choice of action. 
Third, during testing, we do not evaluate information gain at each time step since the DQN agent has learned the mapping between patient state and Q values associated with different actions as a sort of amortized inference. 



\bibliography{nips_2018}

\begin{thebibliography}{29}
\providecommand{\natexlab}[1]{#1}
\providecommand{\url}[1]{\texttt{#1}}
\expandafter\ifx\csname urlstyle\endcsname\relax
  \providecommand{\doi}[1]{doi: #1}\else
  \providecommand{\doi}{doi: \begingroup \urlstyle{rm}\Url}\fi

\bibitem[Ahuja et~al.(2017)Ahuja, Zame, and van~der
  Schaar]{ahuja_dpscreen_2017}
Ahuja, Kartik, Zame, William, and van~der Schaar, Mihaela.
\newblock {DPSCREEN}: {Dynamic} {Personalized} {Screening}.
\newblock In \emph{Advances in {Neural} {Information} {Processing} {Systems}},
  pp.\  1321--1332, 2017.

\bibitem[Contardo et~al.(2016)Contardo, Denoyer, and
  Arti{\`e}res]{contardo2016sequential}
Contardo, Gabriella, Denoyer, Ludovic, and Arti{\`e}res, Thierry.
\newblock Sequential cost-sensitive feature acquisition.
\newblock In \emph{International Symposium on Intelligent Data Analysis}, pp.\
  284--294. Springer, 2016.

\bibitem[Dewan et~al.(2017)Dewan, Galvez, Polsky, Kreher, Kraus, Ahumada,
  McCloskey, and Wolfe]{dewan2017reducing}
Dewan, Maya, Galvez, Jorge, Polsky, Tracey, Kreher, Genna, Kraus, Blair,
  Ahumada, Luis, McCloskey, John, and Wolfe, Heather.
\newblock Reducing unnecessary postoperative complete blood count testing in
  the pediatric intensive care unit.
\newblock \emph{The Permanente journal}, 21, 2017.

\bibitem[Eyster \& Bernene(1973)Eyster and Bernene]{eyster1973nosocomial}
Eyster, Elaine and Bernene, James.
\newblock Nosocomial anemia.
\newblock \emph{JAMA}, 223\penalty0 (1):\penalty0 73--74, 1973.

\bibitem[Ezzie et~al.(2007)Ezzie, Aberegg, and O'Brien]{ezzie2007laboratory}
Ezzie, Michael~E, Aberegg, Scott~K, and O'Brien, James~M.
\newblock Laboratory testing in the intensive care unit.
\newblock \emph{Critical care clinics}, 23\penalty0 (3):\penalty0 435--465,
  2007.

\bibitem[Feldman(2009)]{feldman2009managing}
Feldman, L.
\newblock Managing the cost of diagnosis.
\newblock \emph{Managed care (Langhorne, Pa.)}, 18\penalty0 (5):\penalty0 43,
  2009.

\bibitem[Futoma et~al.(2017)Futoma, Hariharan, Sendak, Brajer, Clement, Bedoya,
  O'Brien, and Heller]{futoma_improved_2017}
Futoma, Joseph, Hariharan, Sanjay, Sendak, Mark, Brajer, Nathan, Clement,
  Meredith, Bedoya, Armando, O'Brien, Cara, and Heller, Katherine.
\newblock An {Improved} {Multi}-{Output} {Gaussian} {Process} {RNN} with
  {Real}-{Time} {Validation} for {Early} {Sepsis} {Detection}.
\newblock \emph{arXiv:1708.05894 [stat]}, August 2017.
\newblock arXiv: 1708.05894.

\bibitem[Futoma et~al.(2018)Futoma, Lin, Sendak, Bedoya, Clement, O'Brien, and
  Heller]{futoma2018learning}
Futoma, Joseph, Lin, Anthony, Sendak, Mark, Bedoya, Armando, Clement, Meredith,
  O'Brien, Cara, and Heller, Katherine.
\newblock Learning to treat sepsis with multi-output gaussian process deep
  recurrent q-networks, 2018.
\newblock URL \url{https://openreview.net/forum?id=SyxCqGbRZ}.

\bibitem[Ghassemi et~al.(2015)Ghassemi, Pimentel, Naumann, Brennan, Clifton,
  Szolovits, and Feng]{ghassemi2015multivariate}
Ghassemi, Marzyeh, Pimentel, Marco~AF, Naumann, Tristan, Brennan, Thomas,
  Clifton, David~A, Szolovits, Peter, and Feng, Mengling.
\newblock A multivariate timeseries modeling approach to severity of illness
  assessment and forecasting in icu with sparse, heterogeneous clinical data.
\newblock 2015.

\bibitem[Harutyunyan et~al.(2017)Harutyunyan, Khachatrian, Kale, and
  Galstyan]{harutyunyan2017multitask}
Harutyunyan, Hrayr, Khachatrian, Hrant, Kale, David~C, and Galstyan, Aram.
\newblock Multitask learning and benchmarking with clinical time series data.
\newblock \emph{arXiv preprint arXiv:1703.07771}, 2017.

\bibitem[He et~al.(2016)He, Mineiro, and Karampatziakis]{he2016active}
He, He, Mineiro, Paul, and Karampatziakis, Nikos.
\newblock Active information acquisition.
\newblock \emph{arXiv preprint arXiv:1602.02181}, 2016.

\bibitem[Hochreiter \& Schmidhuber(1997)Hochreiter and
  Schmidhuber]{hochreiter1997long}
Hochreiter, Sepp and Schmidhuber, J{\"u}rgen.
\newblock Long short-term memory.
\newblock \emph{Neural computation}, 9\penalty0 (8):\penalty0 1735--1780, 1997.

\bibitem[Hoffman \& Cooper(2012)Hoffman and Cooper]{hoffman2012overdiagnosis}
Hoffman, Jerome~R and Cooper, Richelle~J.
\newblock Overdiagnosis of disease: a modern epidemic.
\newblock \emph{Archives of internal medicine}, 172\penalty0 (15):\penalty0
  1123--1124, 2012.

\bibitem[Hwang et~al.(2017)Hwang, Choi, and Yoon]{hwang2017disease}
Hwang, Uiwon, Choi, Sungwoon, and Yoon, Sungroh.
\newblock Disease prediction from electronic health records using generative
  adversarial networks.
\newblock \emph{arXiv preprint arXiv:1711.04126}, 2017.

\bibitem[Iosfina et~al.(2013)Iosfina, Merkeley, Cessford, Geller, Amiri,
  Baradaran, Norena, Ayas, and Dodek]{iosfina2013implementation}
Iosfina, Ioulia, Merkeley, Hayley, Cessford, Tara, Geller, Georgia, Amiri,
  Neda, Baradaran, Nazli, Norena, Monica, Ayas, Najib, and Dodek, Peter~M.
\newblock Implementation of an on-demand strategy for routine blood testing in
  icu patients.
\newblock In \emph{D23. QUALITY IMPROVEMENT IN CRITICAL CARE}, pp.\
  A5322--A5322. Am Thoracic Soc, 2013.

\bibitem[Jiang \& Li(2015)Jiang and Li]{jiang2015doubly}
Jiang, Nan and Li, Lihong.
\newblock Doubly robust off-policy value evaluation for reinforcement learning.
\newblock \emph{arXiv preprint arXiv:1511.03722}, 2015.

\bibitem[Johnson et~al.(2016)Johnson, Pollard, Shen, Li-wei, Feng, Ghassemi,
  Moody, Szolovits, Celi, and Mark]{johnson2016mimic}
Johnson, Alistair~EW, Pollard, Tom~J, Shen, Lu, Li-wei, H~Lehman, Feng,
  Mengling, Ghassemi, Mohammad, Moody, Benjamin, Szolovits, Peter, Celi,
  Leo~Anthony, and Mark, Roger~G.
\newblock Mimic-iii, a freely accessible critical care database.
\newblock \emph{Scientific data}, 3:\penalty0 160035, 2016.

\bibitem[Kotecha et~al.(2017)Kotecha, Shapiro, Cardasis, and
  Narayanswami]{kotecha2017reducing}
Kotecha, Nisha, Shapiro, Janet~M, Cardasis, John, and Narayanswami, Gopal.
\newblock Reducing unnecessary laboratory testing in the medical icu.
\newblock \emph{The American journal of medicine}, 130\penalty0 (6):\penalty0
  648--651, 2017.

\bibitem[Lipton et~al.()Lipton, Kale, and Wetzel]{lipton2016modeling}
Lipton, Zachary~C, Kale, David~C, and Wetzel, Randall.
\newblock Modeling missing data in clinical time series with rnns.

\bibitem[Mayr et~al.(2014)Mayr, Yende, and Angus]{mayr2014epidemiology}
Mayr, Florian~B, Yende, Sachin, and Angus, Derek~C.
\newblock Epidemiology of severe sepsis.
\newblock \emph{Virulence}, 5\penalty0 (1):\penalty0 4--11, 2014.

\bibitem[Moynihan et~al.(2012)Moynihan, Doust, and
  Henry]{moynihan2012preventing}
Moynihan, Ray, Doust, Jenny, and Henry, David.
\newblock Preventing overdiagnosis: how to stop harming the healthy.
\newblock \emph{BMJ: British Medical Journal (Online)}, 344, 2012.

\bibitem[Pageler et~al.(2013)Pageler, Franzon, Longhurst, Wood, Shin, Adams,
  Widen, and Cornfield]{pageler2013embedding}
Pageler, Natalie~M, Franzon, Deborah, Longhurst, Christopher~A, Wood, Matthew,
  Shin, Andrew~Y, Adams, Eloa~S, Widen, Eric, and Cornfield, David~N.
\newblock Embedding time-limited laboratory orders within computerized provider
  order entry reduces laboratory utilization.
\newblock \emph{Pediatric Critical Care Medicine}, 14\penalty0 (4):\penalty0
  413--419, 2013.

\bibitem[Raghu et~al.(2017)Raghu, Komorowski, Celi, Szolovits, and
  Ghassemi]{raghu2017continuous}
Raghu, Aniruddh, Komorowski, Matthieu, Celi, Leo~Anthony, Szolovits, Peter, and
  Ghassemi, Marzyeh.
\newblock Continuous state-space models for optimal sepsis treatment-a deep
  reinforcement learning approach.
\newblock \emph{arXiv preprint arXiv:1705.08422}, 2017.

\bibitem[Shim et~al.(2017)Shim, Hwang, and Yang]{shim2017pay}
Shim, Hajin, Hwang, Sung~Ju, and Yang, Eunho.
\newblock Why pay more when you can pay less: A joint learning framework for
  active feature acquisition and classification.
\newblock \emph{arXiv preprint arXiv:1709.05964}, 2017.

\bibitem[Soleimani et~al.(2017)Soleimani, Hensman, and
  Saria]{soleimani_scalable_2017}
Soleimani, Hossein, Hensman, James, and Saria, Suchi.
\newblock Scalable {Joint} {Models} for {Reliable} {Uncertainty}-{Aware}
  {Event} {Prediction}.
\newblock \emph{arXiv:1708.04757 [cs, stat]}, August 2017.
\newblock URL \url{http://arxiv.org/abs/1708.04757}.
\newblock 00000 arXiv: 1708.04757.

\bibitem[Wang et~al.(2015)Wang, Schaul, Hessel, Van~Hasselt, Lanctot, and
  De~Freitas]{wang2015dueling}
Wang, Ziyu, Schaul, Tom, Hessel, Matteo, Van~Hasselt, Hado, Lanctot, Marc, and
  De~Freitas, Nando.
\newblock Dueling network architectures for deep reinforcement learning.
\newblock \emph{arXiv preprint arXiv:1511.06581}, 2015.

\bibitem[Weng et~al.(2017)Weng, Gao, He, Yan, and
  Szolovits]{weng2017representation}
Weng, Wei-Hung, Gao, Mingwu, He, Ze, Yan, Susu, and Szolovits, Peter.
\newblock Representation and reinforcement learning for personalized glycemic
  control in septic patients.
\newblock \emph{arXiv preprint arXiv:1712.00654}, 2017.

\bibitem[Yoon et~al.(2018)Yoon, Zame, and van~der Schaar]{yoon2018deep}
Yoon, Jinsung, Zame, William~R, and van~der Schaar, Mihaela.
\newblock Deep sensing: Active sensing using multi-directional recurrent neural
  networks.
\newblock 2018.

\bibitem[Zhang et~al.(2017)Zhang, Xie, Wang, and Xing]{zhang2017medical}
Zhang, Shiyue, Xie, Pengtao, Wang, Dong, and Xing, Eric~P.
\newblock Medical diagnosis from laboratory tests by combining generative and
  discriminative learning.
\newblock \emph{arXiv preprint arXiv:1711.04329}, 2017.

\end{thebibliography}
\bibliographystyle{icml2017}

\newpage
\appendix

\section{MIMIC3 Preprocessing for Survival Forecasting}
\label{appendix:preprocessing}
We use the publicly available dataset MIMIC3 \citep{johnson2016mimic} and then follow the preprocessing of \citet{harutyunyan2017multitask} for the in-hospital mortality prediction task. It excludes the neonatal and pediatric patients 
and patients with multiple ICU stays. The training set consists of $35,725$ patients with $10.81\%$ mortality rate, and test set has $6,294$ patients with $9.94\%$ mortality rate.
We then split $15\%$ of our training set as our validation set. 



For classifier training, we uniformly take $6$ timepoints within the last $24$ hours of each patient trajectory. For each prediction point, we set the label as $1$ if the time is within $12$ hours before death and $0$ otherwise.
We only include the prediction points with at least $3$ hours of history and $5$ measurement values.
For RL, we take the last $24$ hours of each dying patient and discretize it into $15$ minutes interval. We only include the patients with at least $12$ hours to remove unstable trajectories.

We select $34$ time-series features that have low missingness and are known to be correlated with sepsis, which is well-known to be the cause of death in ICU \citep{mayr2014epidemiology}. Other $8$ static demographic features such as race or gender are included as well. We show the full lists of features in the appendix \ref{appendix:features}.

\section{Feature choices in MIMIC3}
\label{appendix:features}
We use the following set of $8$ demographic features: age, gender and one-hot encoding of $5$ ethnicity.

We use the following set of $34$ time-series features: 
Albumin, Bicarbonate, Bilirubin, Blood urea nitrogen, CO2, Calcium, Calcium ionized, Capillary refill rate, Chloride, Cholesterol, Creatinine, Diastolic blood pressure, Fraction inspired oxygen, Glasgow coma scale total, Glucose, Heart Rate, Hemoglobin, Lactate, Magnesium, Mean blood pressure, Oxygen saturation, Partial pressure of carbon dioxide, Partial pressure of oxygen, Partial thromboplastin time, Platelets, Potassium, Prothrombin time, Respiratory rate, Systolic blood pressure, Temperature, Urine output, Weight, White blood cell count, and pH.

\section{Classifier Training Details and Performances}
\label{appendix:classifier_performance}
We train the RNN as follows.
We used LSTM with $1$ hidden layer of $32$ nodes. We regularize the neural network with $\lambda=1$e$-5$ as $\ell_2$ regularization and dropout rate of $0.3$ for input and output layer, and $0.5$ for the hidden layer. 
We use mean imputation (set the missing value as the feature mean) for the time-series features, and add missingness indicators for each feature \citep{lipton2016modeling}.
We discretize the time series into $1$ hour interval and take mean value if there are multiple measurements per interval. The RNN takes these 1-hour discretized grid point for up to $24$ hours time point to classify.
We train two other baselines: Logistic Regression (LR) with $\lambda=1$e$-5$ as $\ell_2$ regularization and Random Forest (RF) with $500$ trees. We concatenate all the features across all the time points.

We show the classifier's test set performance in Table \ref{table:classifier_performance} with the prediction horizon as $12$ hours. 
We show that RNN is the best among all the classifiers.

\begin{table}[ht]
  \centering 
  \caption{The test set performances of the trained classifiers in $12$ hours mortality prediction.} 
\begin{tabular}{|l|l|l|}
  \toprule
& AUC & AUPR \\ \midrule
LR & $0.894$ & $0.514$ \\
RF & $0.920$ & $0.586$ \\ \midrule
RNN & $\bm{0.931}$ & $\bm{0.628}$ \\
\bottomrule
\end{tabular}
\label{table:classifier_performance}
\end{table}

\end{document}